\def\year{2022}\relax
\documentclass[letterpaper]{article} % DO NOT CHANGE THIS
\usepackage{aaai22}  % DO NOT CHANGE THIS
\usepackage{times}  % DO NOT CHANGE THIS
\usepackage{helvet}  % DO NOT CHANGE THIS
\usepackage{courier}  % DO NOT CHANGE THIS
\usepackage[hyphens]{url}  % DO NOT CHANGE THIS
\usepackage{graphicx} % DO NOT CHANGE THIS
\usepackage{natbib}  % DO NOT CHANGE THIS AND DO NOT ADD ANY OPTIONS TO IT
\usepackage{caption} % DO NOT CHANGE THIS AND DO NOT ADD ANY OPTIONS TO IT
\DeclareCaptionStyle{ruled}{labelfont=normalfont,labelsep=colon,strut=off} % DO NOT CHANGE THIS
\usepackage{algorithm}
\usepackage{algorithmic}

\usepackage{newfloat}
\usepackage{listings}
\floatstyle{ruled}
\newfloat{listing}{tb}{lst}{}
\floatname{listing}{Listing}
\usepackage{amssymb}
\usepackage{booktabs}
\usepackage{makecell}
\usepackage{multirow}

\title{Sparse Structure Learning via Graph Neural Networks\\for Inductive Document Classification}
\author{
    Yinhua Piao\textsuperscript{\rm 1},
    Sangseon Lee\textsuperscript{\rm 2},
    Dohoon Lee\textsuperscript{\rm 3},
    Sun Kim\textsuperscript{\rm 1,3,4,5}
}
\affiliations{
    \textsuperscript{\rm 1} Department of Computer Science and Engineering, Seoul National University\\
    \textsuperscript{\rm 2} Institute of Computer Technology, Seoul National University\\
    \textsuperscript{\rm 3} Bioinformatics Institute, Seoul National University, \textsuperscript{\rm 4} AIGENDRUG Co., Ltd.\\
    \textsuperscript{\rm 5} Interdisciplinary Program in Artificial Intelligence, Seoul National University\\
    \{2018-27910, sangseon486, apap7, sunkim.bioinfo\}@snu.ac.kr 

}

\usepackage{bibentry}
\begin{document}

\maketitle

\begin{abstract}
Recently, graph neural networks (GNNs) have been widely used for document classification. However, most existing methods are based on static word co-occurrence graphs without sentence-level information, which poses three challenges:(1) word ambiguity, (2) word synonymity, and (3) dynamic contextual dependency. To address these challenges, we propose a novel GNN-based sparse structure learning model for inductive document classification. Specifically, a document-level graph is initially generated by a disjoint union of sentence-level word co-occurrence graphs. Our model collects a set of trainable edges connecting disjoint words between sentences, and employs structure learning to sparsely select edges with dynamic contextual dependencies. Graphs with sparse structures can jointly exploit local and global contextual information in documents through GNNs. For inductive learning, the refined document graph is further fed into a general readout function for graph-level classification and optimization in an end-to-end manner. Extensive experiments on several real-world datasets demonstrate that the proposed model outperforms most state-of-the-art results, and reveal the necessity to learn sparse structures for each document.
\end{abstract}

\section{Introduction}
Document classification, a task of using algorithms to automatically classify the input document to one or multiple categories, is one of the most fundamental tasks in the field of Natural Language Processing (NLP).
The essential part of document classification is to extract features that can represent the documents. 
Conventional approaches use hand-crafted features, e.g., bag-of-words, term frequency-inverse document frequency.
With the advent of deep learning technologies, related works such as Word2Vec \cite{mikolov2013distributed}, utilize contextual information to learn word representations.
Considering the word order in a sequence, many models adopt sequence-based models including recurrent neural networks (RNNs) \cite{mikolov2010recurrent, tai2015improved, liu2016recurrent}, and convolutional neural networks (CNNs) \cite{kim-2014-convolutional, zhang2015character}.
Although these methods can capture the local contextual information in the document, sequence-based models still have difficulty in capturing long-range word co-occurrence information.
With rapid adoption of graph neural networks (GNNs) \cite{kipf2017semi}, GNNs can be designed to capture non-consecutive word dependencies in the document. Therefore, GNNs have recently been used for document classification. 
TextGCN \cite{yao2019graph} first applies GNNs on one corpus graph for node-level document classification task.
Huang et al. \cite{huang2019text} transform TextGCN to graph-level prediction to reduce the memory consumption during training.
To improve generalization performance for new documents, there are also works for inductive document classification. TextING \cite{zhang2020every} constructs an individual graph for each document where local word interactions can be learned. HyperGAT \cite{ding2020more} improves expressive power of inductive model by exploiting high-order relations in document-level hypergraphs.
These methods, with satisfying results, can empirically prove that graph-based models can indeed capture long-range word dependencies which benefits the model performance.

Nevertheless, almost all graph-based methods are designed to construct static word co-occurrence graph for the whole document without considering sentence-level information.
Each unique word in the graph is mapped to only one representation in the latent space, which may bring three potential challenges: (1) \textit{Word ambiguity}. In the real world, most words may have multiple meanings, and in different contexts, a single word may have completely different meanings depending on the context.
In the static graph, an anchor word with multiple completely different meanings is connected to all adjacent words as \textit{1-hop} neighbors, which can misguide GNNs to blindly combine global information and confuse syntactic information, as well as degrade local information.
(2) \textit{Word synonymity}. Non-consecutive words in the static graph can be mapped to similar positions in the latent space, in many cases owing to their proximity to the same anchor word.
However, most words have their synonyms, and words adjacent to synonyms should be mapped similarly. Therefore, some long-range information between synonyms may still not be captured in a static word co-occurrence graph.
Last, (3) \textit{Dynamic contextual dependency}.
Most GNN-based approaches consider nodes and their neighbors to be homogeneous in the static document graph, allowing simultaneous layer-by-layer message passing.
However, syntactic and semantic information should be passed specifically and develop dynamically, rather than at each hierarchy simultaneously.
In conclusion, it is necessary to learn the dynamic graph structure of documents with local syntactic and global semantic information with dynamic contextual dependency.

To address the gaps of aforementioned limitations, we propose a novel sparse graph structure learning for inductive document classification, which constructs learnable and individual graphs for each document.
Specifically, nodes in the document graph first pass messages to their intra-sentence neighbors and inter-sentence neighbors, which can be regarded as local syntactic messages and global semantic messages, respectively. Then we apply the proposed sparse structure learning with Gumbel-softmax trick to learn and update the graph structure, aiming for dynamic contextual dependencies with fewer noise from layer to layer. The learned graph with local and global information is further fed into a general readout function for classification and optimization in an end-to-end manner.
The contribution of this paper is summarized as follows and all the code
publicly available at \texttt{https://github.com/qkrdmsghk/TextSSL} :
\begin{itemize}
\item We construct a trainable individual graph consisting of sentence-level subgraphs for each document. To our best knowledge, we are the first to construct a trainable graph for inductive document classification. 
\item We propose a sparse structure learning model via GNNs to learn an effective and efficient structure with dynamic syntactic and semantic information for each document. 
\item We conduct extensive empirical experiments on several real-world. In the experiments, our model outperforms most existing approaches, which supports the effectiveness of our approach.
\end{itemize}

\section{Preliminaries}
\label{sec:pre}
\subsection{Graph Neural Networks}
GNNs use the graph structure and node features to learn representation vectors for each node in the graph to conduct the node-level prediction task, or combine all of them to predict property of the graph. 
Recent studies have focused on spatial-based GNNs that describe a general framework of message passing networks. 
The essence of the message passing network is to iteratively propagate and aggregate information across the nodes of the graph. Formally, the $k$-th iteration of message passing process in a GNN consisting of aggregation operation and update operation that is defined as :
\begin{equation}
    h_v^{k} = \phi \left(f^{(k)}(h_v^{(k-1)}, \{ h_u^{(k-1)}: u \in \mathcal{N}_v \})\right),
\end{equation}
where $h^k_v$ denotes the embedding vector at layer $k$ associated with node $v$, the function $f^{(k)}(\cdot)$ aggregates and updates the node representations from their neighbor nodes at the previous layer. $\phi(\cdot)$ represents an injection function, such as non-linear activation function. 
For graph classification, the readout function aggregates node representations to obtain the entire graph's representation $h_G$:
\begin{equation}
    h_G = R(\{ h_v^{(K)} | v \in G \}).
\end{equation}
where $R(\cdot)$ denotes a simple permutation invariant function such as global average pooling or global max pooling after $K$ iterations. 

\subsection{Gumbel-Softmax Distribution}
Formally, let a discrete variable $\pi$ has a distribution of probabilities (\(\phi_1, ...,\phi_n\)) with class $C=\{c_1,...,c_n\}$.
Gumbel-max \cite{gumbel1954statistical} provides an efficient way for the categorical distribution to sample $x_\pi$ with:
\begin{equation}
x_{\pi}=\mathrm{argmax}(\log\phi_i + G_i)
\end{equation}
where $G_i$ is a Gumbel noise sampled from \textit{Gumbel}(0,1).
To solve non-differentiable problem of Gumbel-Max, \citet{jang2016categorical} propose Gumbel-Softmax to approximate it as follows:
\begin{equation}
\hat{x}_\pi=\frac{\exp((\log(\phi_i)+G_i)/\tau)}{\sum_{j=1}^{n}\exp((\log(\phi_j)+G_j)/\tau)}
\end{equation}
where a softmax function with an adjustable temperature $\tau$ is to control the argmax operation to make it possible for a differentiable optimization.

\begin{figure*}[t]
\centering
\includegraphics[width=1.0\textwidth]{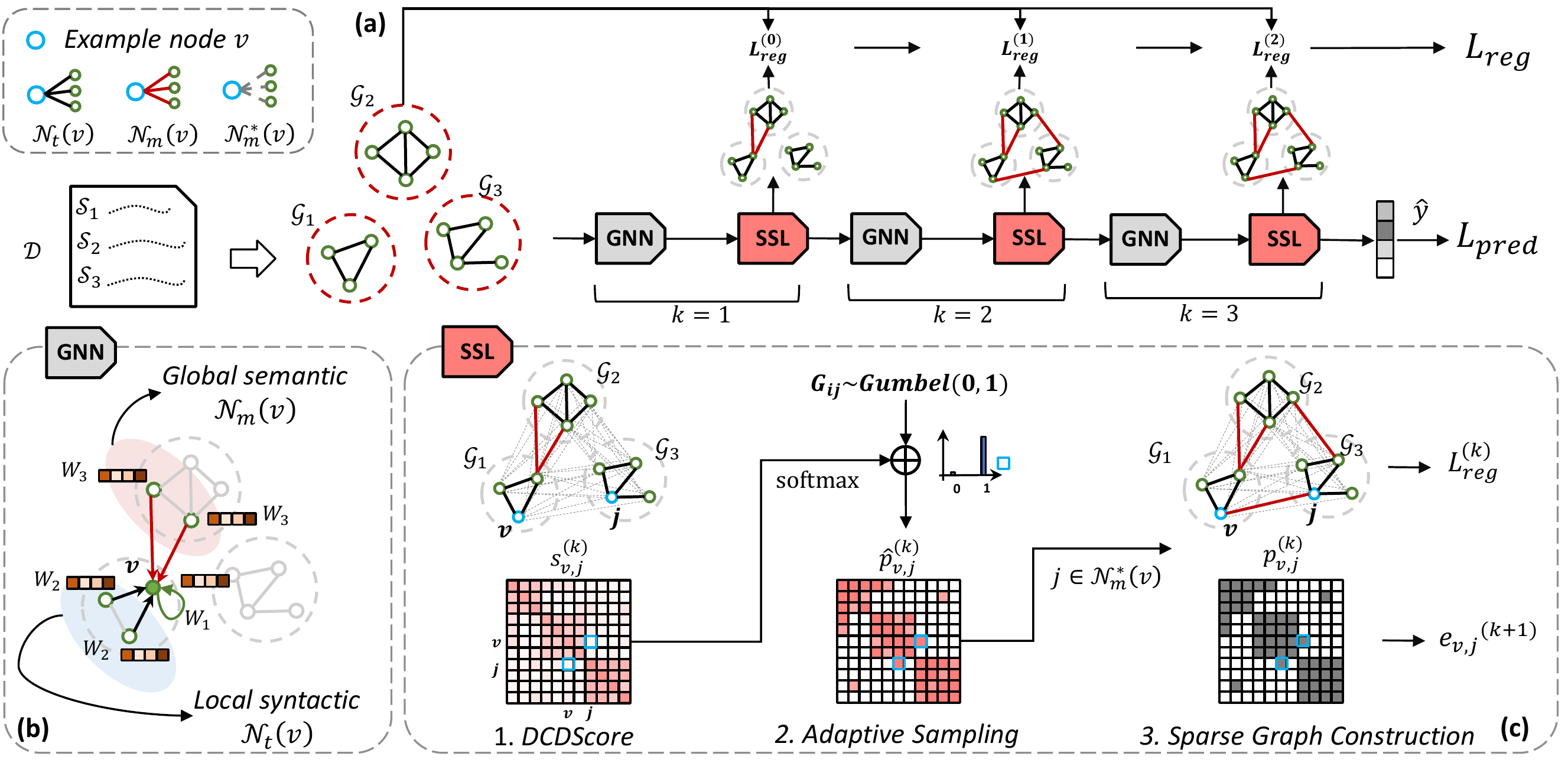} % Reduce the figure size so that it is slightly narrower than the column.
\caption{Overview of the proposed model. (a) Model framework. (b) GNN: Local and Global Joint Message Passing. (c) SSL: Sparse Structure Learning contains (c.1) Dynamic Contextual Dependency Score, (c.2) Adaptive Sampling for Sparse Structure, and (c.3) Reconstructing Sparse Graph. }
\label{model_overview}
\end{figure*}

\section{The Proposed Model}
In this section, we introduce our model for inductive document classification. The proposed model consists of three main parts. We first construct document-level graphs in which the node embedding is learned with a local and global message passing operation. Based on the node embedding, we propose a sparse structure learning for the graph structure refinement. Last, we regularize the graph structure to preserve consistency of the original syntactic information.

\subsection{Problem Definition} We are given a set of documents \(\mathcal{D}\) with a set of document labels \(\mathcal{Y}\) where each document $d\in\mathcal{D}$ may have multiple sentences \(\mathcal{S}^{d}=[s^d_0, ..., s^d_k]\) and each sentence $s^d\in\mathcal{S}^d$ is composed of multiple words \(\mathcal{W}^{s^{d}}=[{w_0}^{s^d}, ..., {w_n}^{s^d}]\). We represent each document as an individual graph \(\mathcal{G}^{d}\) using its hierarchical structure for inductive learning. For simplicity, we omit the document index $d$ throughout the paper.
The document graph is composed of multiple words in $\mathcal{W}^s$ and their connections. Our goal is to learn the structure and make a prediction for each \(\mathcal{G}\).

\subsection{Graph Construction}

\subsubsection{Definition 1.}\textbf{Sentence-level Subgraph} Given a sentence \(s_i \in \mathcal{S} \), a sentence-level subgraph \( \mathcal{G}_i = (\mathcal{V}_i, \mathcal{E}_i)\) can represent the sentence $s_i$ as a word co-occurrence graph. 
The node set \(\mathcal{V}_i\) contains words in sentence $s_i$. The edge set \(\mathcal{E}_i\) contains all connections between any pair of words in \(\mathcal{V}_i\) that can co-occur in the same fixed-size sliding window \cite{mihalcea2004textrank}. Therefore, a preliminary document graph \(\tilde{\mathcal{G}}=(\mathcal{V}, \mathcal{E})\) can be represented by taking the disjoint union of all sentence-level subgraphs \( \mathcal{G}_\mathcal{S}=\{ {\mathcal{G}_1,..., \mathcal{G}_n } \}\), where $n$ denotes the number of sentences within the document.

\subsubsection{Definition 2.}\textbf{Local Syntactic Neighbor} Given a node \(v \in \mathcal{V}\) in a preliminary document graph \(\tilde{\mathcal{G}}\), we define a local syntactic neighbor \(u \in \mathcal{N}_t(v)\) that is adjacent to node \(v\) within sentence-level subgraphs \(\mathcal{G}_\mathcal{S}\). Since sentence-level subgraphs \(\mathcal{G}_\mathcal{S}\) contain relatively more invariant and syntactic information, we constrain the local syntactic neighbors to be static and deterministic during graph structure learning.
\subsubsection{Definition 3.}\textbf{Global Semantic Neighbor} Given a node \(v \in \mathcal{V}\) in a preliminary document graph \(\tilde{\mathcal{G}}\), we define a global semantic neighbor \(z \in \mathcal{N}_m(v)\) that can have dynamic relation with node \(v\) between sentence-level subgraphs \(\mathcal{G}_s\). Global semantic neighbors of each node in \(\mathcal{V}\) are dynamic and can be learned and selected via structure learning. 
\subsubsection{A document-level graph} \(\mathcal{G}=(\mathcal{V},  \{\mathcal{E}_t\cup{\mathcal{E}_m\}})\) is finally composed of all sentence-level subgraphs \(\mathcal{G}_{\mathcal{S}}\), where edges \(\mathcal{E}_t\) connect nodes and their local syntactic neighbors \(\mathcal{N}_t(\cdot)\) and edges \(\mathcal{E}_m\) connect nodes and their global semantic neighbors \(\mathcal{N}_m(\cdot)\). In sparse graph structure learning module, global semantic neighbors can be learned and chosen dynamically during which local syntactic edges can guide the dynamic edge relaxation. 

\subsection{Local and Global Joint Message Passing }
Unlike existing GNNs considering that all nodes are homogeneous, we differently aggregate the neighbor messages by distinguishing the neighbor node types (local syntactic neighbor and global semantic neighbor) to update the node representation. The message passing part can be reformulated as:
\begin{equation}
h_{v}^{(k)}=\phi\left(h_{v}^{(k-1)}\mathbf{W}_1^{(k)}+t_{v}^{(k)}\mathbf{W}_2^{(k)}+m_v^{(k)}\mathbf{W}_3^{(k)}\right),
\end{equation}
where function $\phi$ denotes an injective function ReLU(\(\cdot\)). \(h_{v}^{(k)}\in\mathbb{R}^{b}\) is the node representation vector and $b$ is the number of hidden dimension. The local syntactic neighbor representations $t_{v}^{(k)}\in\mathbb{R}^{b}$ and global semantic neighbor representations $m_{v}^{(k)}\in\mathbb{R}^{b}$ can be expressed as:
\begin{equation}
{t}_{v}^{(k)} =\sum_{u\in{\mathcal{N}_t(v)\cup{\{v\}}}}\frac{e_{u,v}}{\sqrt{\hat{\zeta}_u\hat{\zeta}_v}}h_{u}^{(k-1)}
\end{equation}
\begin{equation}
m_{v}^{(k)}=\sum_{z\in{{\mathcal{N}_m(v)}}^{(k-1)}}\frac{e_{z,v}}{\sqrt{\hat{\zeta}_z\hat{\zeta}_v}}h_{z}^{(k-1)}
\end{equation}
where \(e_{u,v}\in\mathcal{E}_t\) represents edge weight between node $v$ and its local syntactic neighbor $u$. \(e_{z,v}\in\mathcal{E}^{(k-1)}_m\) represents edge weight between node $v$ and its global semantic neighbor $z$.
Here, we normalize the raw edge weight to prevent from the influence of degree bias. Motivated by \citet{kipf2017semi}, we divide both edges $e_{u,v}$ and $e_{z,v}$  by the $\hat{\zeta}_v$ where, $\hat{\zeta}_{v}=\sum_{j\in\mathcal{N}}\hat{A}_{vj}$ with self-looped adjacency matrix $\hat{A}=A+I$.

With the iteration of local and global message passing operation on $\mathcal{G}$, nodes combine high-order neighbor nodes within sentences $\mathcal{G}_\mathcal{S}$ and also select and combine dynamic neighbor nodes between sentences. In this way, the local syntactic information can be combined with global semantic information in each layer, which can learn the rich contextual information in the document. 

\subsection{Sparse Structure Learning}

Since relationships among sentences are not known prior, it is vital to refine the document-level graph $\mathcal{G}$ by exploiting local and global contextual dependencies. One feasible approach is learning from a  \textit{Complete Graph} $\mathcal{G^*}=(\mathcal{V},\{\mathcal{E}_t \cup{\mathcal{E}_m^*}\})$ where nodes between subgraphs $\mathcal{G}_\mathcal{S}$ are fully connected. However, fully connected words between sentences usually bring both necessary and unnecessary information, a form of noisy information.
Therefore, in this section, we perform a sparse structure learning, which can be divided into two parts.
(1) Calculating the relative score of global semantic candidate neighbors of each node, which has dynamic contextual dependency.
(2) Conducting adaptive hard selection for global semantic candidate neighbors using Gumbel-softmax approach. Finally, global semantic neighbor set is updated and the document-level graph $\mathcal{G}$ can obtain a sparse structure from $\mathcal{G}^*$.

\subsubsection{Dynamic Contextual Dependency Score}
Given a node $v \in \mathcal{V}$ in a complete graph $\mathcal{G}^*$,  all neighbors of node $v$ are in $\mathcal{N}^*(v)$, where we can obtain  $\mathcal{N}_m^*(v)=\mathcal{N}^*(v)-\mathcal{N}(v)^{(k-1)}$ that contains all global semantic candidate neighbors of node $v$. We first calculate \textit{attention coefficient score} between each neighbor $j\in\mathcal{N}^*(v)$ and node $v$ as follows:
\begin{equation}
a_{v,j}^{*(k)} = \psi\left({\mathbf{a}^{(k)}}^{\top}[h_{v}^{(k)}\mathbf{W}^{(k)}||h_{j}^{(k)}\mathbf{W}^{(k)}]\right)
\end{equation}
where $\mathbf{W}^{(k)}\in\mathbb{R}^{b\times{b}}$ denotes the projection for node features $h_v\in\mathbb{R}^{1\times{b}}$ and $h_j\in\mathbb{R}^{n\times{b}}$. $k$ denotes the current layer of our model. We adopt function $\psi$ as LeakyReLU($\cdot$) activation function, and $\mathbf{a}\in\mathbb{R}^{b\times{1}}$ is a learnable vector.
To consider the correlation between current local and global contextual information, we normalize $a_{v,j}^{*(k)}$ with softmax function across the nodes to calculate dynamic context dependency score:
\begin{equation}
s_{v,j}^{(k)} = \frac{\exp(a_{v,j}^{*(k)})}{\sum_{u\in\mathcal{N}^*(v)}{\exp(a_{v,u}^{*(k)})}}.
\end{equation}
We adopt normalization operation on $\mathcal{N}^*(v)$ which both contains: existing neighbors $\mathcal{N}(v)^{(k-1)}$ and the global semantic candidate neighbors $\mathcal{N}^*_m(v)$ that are going to be selected in current layer. The existing neighbors consist of both local syntactic neighbors $\mathcal{N}_t(v)$ and global semantic neighbors $\mathcal{N}_m(v)^{(k-1)}$ that are selected in previous layers. Therefore the score $s_{v,j}^{(k)}$ of global semantic candidate neighbors can elucidate a relative difference compared to existing dynamic contextual dependency of node $v$.
\subsubsection{Sampling Adaptive Neighbors for Sparse Structure}
Based on the dynamic contextual dependency score $s_{v,j}^{(k)}$, we perform an adaptive sampling on the $\mathcal{G}^*$ . To determine sparse edge, we set a threshold to select meaningful global semantic neighbors $j$ for each node $v$ via argmax operation. However, this operation is not be differentiable during back-propagation process to optimize model. Inspired from \citet{jang2016categorical}, we first generate a neighbor selector $p_{v,j}^{(k)}\in\{0, 1\}$ from the Bernoulli distribution $\{\pi_1:=s_{v,j}^{(k)}, \pi_0:=1-s_{v,j}^{(k)}\}$ and adopt Gumbel-Softmax approach to generate differentiable probability $\hat{p}_{v,j}^{(k)}$ of selector samples ${p}_{v,j}^{(k)}$ as follows:
\begin{equation}
\hat{p}_{v,j}^{(k)} = \frac{\exp((\log{\pi_1}+g_1)/\tau)}{\sum_{i\in\{0,1\}}\exp((\log{\pi_i}+g_i)/\tau)},
\label{10}
\end{equation}
where $g_1$ and $g_0$ are i.i.d variables sampled from Gumbel distribution, and $\tau\in(0, \infty)$ denotes temperature parameter. As $\tau\to{0}$, $\hat{p}_{v,j}^{(k)}$ can be annealed to categorical distribution. Then we can get a discrete neighbor selector $p_{v,j}^{(k)}$ by setting a threshold $T$.
\subsubsection{Reconstructing Sparse Graph} 
Thus, we can select informative neighbors for node $v$ using $p_{v,j}^{(k)}$. It is worth mentioning that despite $p_{v,j}^{(k)}$ is obtained from $s_{v,j}^{(k)}$ that is calculated by both existing 
neighbors and global semantic candidate neighbors, we only select the neighbors from the candidate neighbor set to maintain the local syntactic topology of the document graph. Specifically, we update the global semantic neighbors $\mathcal{N}_m(v)^{(k)}$ for node $v$ with selected candidate neighbors as follows:
\begin{equation}
\mathcal{N}_{m}(v)^{(k)}=\mathcal{N}_{m}(v)^{(k-1)}\cup{\{j\left[|\right]\forall{j\to{p}_{v,j}^{(k)}=1\}}}.
\end{equation}
where $j\in\mathcal{N}^*_m(v)$. In addition, for static local syntactic neighbors $\mathcal{N}_t(v)$, we compute the entropy to preserve consistency of the original syntactic information and prevent too much structure variation in the graph.
\begin{equation}
L_{reg}^{(k)} = \sum_{v\in\mathcal{V}}{\sum_{j\in\mathcal{N}_t(v)}{- \hat{p}_{v,j}^{(k)}\log{(\hat{p}_{v,j}^{(k)}}})},
\end{equation}
At the last iteration, all of the nodes in the graph $\mathcal{G}$ are fed into readout function with simple summation operation and linear operation. We use cross entropy loss function $l(\cdot,\cdot)$ to measure prediction and true label $y$ of the document.
\begin{equation}
L_{pred} = l(R(h_{v}), y),
\end{equation}
We therefore minimize the loss by summing prediction loss $L_{pred}$ and averaged regularization loss $\lambda\sum_{k}{L_{reg}^{(k)}}$ for each document classification tasks, where $\lambda$ is a hyperparameter to adjust the trade-offs between newly learned structure and original structure.

\begin{table*}[t]
\centering
\begin{tabular}{l|cccrrrrrr}
\toprule
Dataset & \#Docs & \#Training & \#Test & \#Classes ($\rho$) & \#Vocab. & Avg.\#Length & Avg.\#Sentence & \#Prop.NW\\ \midrule\midrule
MR      & 10,662 & 7,108 & 3,554 & 2 (1.0)  & 18,764  & 20.39  & 1.17 & 30.09\%   \\
R8      & 7,674   & 5,485     & 2,189 & 8 (84.7)  & 7,688   & 65.72 &  4.03 & 2.60\%  \\
R52     & 9,100     & 6,532 & 2,568 & 52 (1666.7) & 8,892   & 69.82   &  4.34 & 2.63\% \\
Ohsumed & 7,400       & 3,357     & 4,034     & 23 (62.5) & 14,157  & 135.82 & 8.59 & 8.46\%  \\
20NG    & 18,846   & 11,314    & 7,532     & 20 (1.6) & 42,757  & 221.26 & 6.06 & 7.40\%   \\ \bottomrule
\end{tabular}
%}
\caption{Statistics  of the datasets. $\rho$ denotes class imbalance ratio (the sample size of the most frequent class divided by that of the least frequent class). The Avg.\#Length and the Avg.\#Sentence mean the number of words and the number of sentences in a document, respectively. The \#Prop.NW denotes the proportion of new words in test.}
\label{table1}
\end{table*}

\begin{table*}[t]
\centering
\begin{tabular}{c|ccccccc}
\toprule
\textbf{Categories} & \textbf{Baselines} & \textbf{MR} & \textbf{R8} & \textbf{R52} & \textbf{Ohsumed} & \textbf{20NG}\\ \midrule\midrule
\multirow{2}*{Word-based} 
& fastText           & 72.17$\pm$1.30  & 86.04$\pm$0.24 & 71.55$\pm$0.42 & 14.59$\pm$0.00 & 11.38$\pm$1.18 \\
~ & SWEN             & 76.65$\pm$0.63  & 95.32$\pm$0.26 & 92.94$\pm$0.24 & 63.12$\pm$0.55 & 85.16$\pm$0.29 \\
\midrule
\multirow{3}*{Sentence-based} 
&  CNN-non-static    & 77.75$\pm$0.72 & 95.71$\pm$0.52 & 87.59$\pm$0.48 & 58.44$\pm$1.06 & 82.15$\pm$0.52 \\
~ &  LSTM (pretrain)  & 77.33$\pm$0.89 & 96.09$\pm$0.19 & 90.48$\pm$0.86 & 51.10$\pm$1.50 & 75.43$\pm$1.72 \\
~ &  Bi-LSTM         & 77.68$\pm$0.86 & 96.31$\pm$0.33 & 90.54$\pm$0.91 & 49.27$\pm$1.07 & 73.18$\pm$1.85 \\
\midrule
\multirow{4}*{Graph-based (Tr)} 
& TextGCN           & 76.74$\pm$0.20  & 97.07$\pm$0.10 & 93.56$\pm$0.18 & 68.36$\pm$0.56 & 86.34$\pm$0.09 \\
~ &  Huang et al.   & -  & 97.80$\pm$0.20 & 94.60$\pm$0.30 & 69.40$\pm$0.60 & - \\ 
~ & TensorGCN       & 77.91$\pm$0.07  & \textbf{98.04$\pm$0.08} & 95.05$\pm$0.11 & 70.11$\pm$0.24 & \textbf{87.74$\pm$0.05} \\
~ & DHTG        & 77.21$\pm$0.11  & 97.33$\pm$0.06 & 93.93$\pm$0.10 & 68.80$\pm$0.33 & 87.13$\pm$0.07 \\
\midrule
\multirow{3}*{Graph-based (Ind)} 
% ~ & TextING         & \textbf{79.82$\pm$0.20}  & \textbf{98.04$\pm$0.25} & \textbf{95.48$\pm$0.19} & \textbf{70.42$\pm$0.39} & -            \\
~ & TextING    & 78.93$\pm$0.65  & 97.34$\pm$0.25 & 93.73$\pm$0.47 & 67.95$\pm$0.52 & OOM            \\
% & HyperGAT          & 78.32$\pm$0.27  & 97.97$\pm$0.23 & 94.98$\pm$0.27 & 69.90$\pm$0.34 & 86.62$\pm$0.16 \\
& HyperGAT       & 77.36$\pm$0.22           & 96.82$\pm$0.21 & 94.15$\pm$0.18 & 66.39$\pm$0.65 & 84.65$\pm$0.31 \\
\cmidrule{2-7}
~ & Our proposal    & \textbf{79.74$\pm$0.19} & 97.81$\pm$0.14 & \textbf{95.48$\pm$0.26} & \textbf{70.59$\pm$0.38} & 85.26$\pm$0.28 \\
\bottomrule
\end{tabular}
\caption{Test accuracies of various models on five benchmark datasets. The mean $\pm$ standard deviation of all models are reported an average of 10 executions of each model.
% [EW: according to 10 times run. --> an average of 10 excutions of each model]
Graph-based (Tr) means transductive graph-based methods and Graph-based (Ind) means inductive graph-based methods.}
\label{table2}
\end{table*}

\section{Experiments}
\subsection{Datasets}
For a fair and comprehensive evaluation, we use the same benchmark datasets that are used in \cite{yao2019graph}. There are five datasets in three different domains, including sentiment analysis, news classification, and topic classification domains. We use MR dataset for binary sentiment analysis with either positive or negative polarity. We use three datasets in news classification. 20NG is a newsgroup file dataset with 20 categories and reasonably balanced. R8 and R52 are two subsets of Reuters 21578 \cite{Reuters} datasets with 8 and 52 categories respectively and both two datasets are extremely imbalanced. Ohsumed is a topic classification dataset consisting of medical abstracts with 23 categories, such as cardiovascular disease. 
\subsection{Experimental Settings}
For quantitative evaluation, we follow the same train/test splits and data preprocessing for MR, Ohsumed and 20NG datasets as \cite{kim-2014-convolutional,yao2019graph}. For R8 and R52 datasets, they are only provided by a preprocessed version without punctuation and do not have explicit sample names. Since we use documents with sentence segmentation information to construct graph, we re-extract the data from original Reuters-21578 dataset. 
More details on the preprocessing of R8 and R52 dataset are provided in Appendix.
In each experiment, we randomly select 10\% documents from the training set to build validation set. 
A statistics of the benchmark dataset is listed in Table \ref{table1}.

% \subsection{Experimental Settings}
% For quantitative evaluation, we follow the same train/test splits and data preprocessing for MR, Ohsumed and 20NG datasets as \cite{kim-2014-convolutional,yao2019graph}. For R8 and R52 datasets, they are only provided by a preprocessed version that lack punctuations and do not have explicit sample names. Since we use documents with sentence segmentation information to construct graph, we re-extract the data from original Reuters-21578 dataset. 
% A statistics of the benchmark dataset is listed in Table \ref{table1}. 
% For hyperparameter settings, we search GNN layers from \{2, 3\}. We set the initial node dimension to 300. We use Adam \cite{kingma2014adam} to optimize the model. For baseline models, we either show the results reported in previous research \cite{yao2019graph} or run the codes provided by the authors using the parameters described in the original papers. 
\subsection{Baseline Methods}
In our experiments, the baselines are divided into three categories: word based methods, sequence based methods and graph based methods. In \textit{word-based} methods, we use fastText \cite{joulin2017bag} and SWEM \cite{shen2018baseline}. 
In \textit{sequence-based} methods, we use CNN \cite{kim-2014-convolutional} with pretrained word embedding, and RNN \cite{liu2016recurrent} with pretrained word embedding and its variant models LTSM from \cite{yao2019graph}. \textit{Graph-based} models for document classification can be categorized into transductive learning and inductive learning. We compare a series of GNN-based transductive models, such as TextGCN \cite{yao2019graph}, TensorGCN  \cite{liu2020tensor}, DHTG \cite{wang2020learning}, Huang et al. \cite{huang2019text}. We also compare with recent published inductive models, such as HyperGAT \cite{ding2020more}, TextING \cite{zhang2020every}. Details of these methods are provided in related works.

\subsection{Parameter Settings}
In this part, we describe the hyperparameter settings for model training. First, we search GNN layers from \{2, 3\}, and batch size is chosen from \{16, 64, 128, 256\}. We set the initial node dimension to 300, and then we search the hidden node dimension from \{96, 256, 512\}. The hyperparameters of all datasets are reported in the Appendix. We use Adam \cite{kingma2015adam} to optimize the model. We use PyTorch \cite{paszke2019pytorch} to implement our architecture. For Texting and HyperGAT baselines, we use the same datasets for fair comparisons. All models are trained on a single NVIDIA GeForce RTX 3080 GPU. For baseline models, we either show the results reported in previous research \cite{yao2019graph} or run the codes provided by the authors using the parameters described in the original papers. More details can be found in the Appendix.
\subsection{Experiment Results}
Table \ref{table2} shows performance comparisons of the different methods on five benchmark datasets. Firstly, most graph-based methods outperform both word-based and sequence-based baselines, indicating that long-range dependencies captured by graph-based models benefit document classification.
Next, we compare our model with transductive and inductive graph-based models, respectively. Overall, our model achieves best results among all inductive learning models, indicating that the sparse graph structure learned from our model using dynamic contextual information has a positive effect on inductive learning.
To summarize, our observations are as follows:
\paragraph{Unseen words.}It is noted in Table \ref{table2} that our model significantly outperforms  in MR dataset. According to the \#Prop.NW in Table \ref{table1}, we can find that there are many unseen words in the test set, which indicates that the sparse structure of the documents learned by our model using inductive learning favors the generalization ability.
\paragraph{Document length.}From Table \ref{table1} and Table \ref{table2}, we find a trend that inductive models perform better on short documents (MR, R8, and R52), while most transductive methods perform relatively well on long documents (Osumed, 20NG).
It seems that long documents own denser structures that can benefit message passing for the transductive methods. For inductive learning, the dense structure introduces additional noise, which makes the learning of the model difficult.
Even so, our model combines syntax and global semantics to learn sparse graphs for documents. Therefore the proposed model outperforms all baselines on Ohsumed dataset and all existing inductive methods on 20NG dataset.
\paragraph{Dynamic contextual dependency.} Most notably, our model and TensorGCN achieve the best performance in the inductive and transductive models, respectively.
TensorGCN, like our model, also takes into account both syntactic and sequential information. This suggests that considering sequential, syntactic and semantic information in these datasets can help a lot in document classification.
Unlike TensorGCN, our model is able to perform inductive learning and can leverage the learned sparse dynamic contextual dependencies from rich structured document to improve generalization performance in more complex classification tasks, e.g., unbalanced (R52) and domain-specific (Ohsumed) datasets.

\subsection{Issues in Constructing Document-level Graph}
In this subsection, we analyze the effectiveness of different ways to construct document-level graphs for the document classification task: (1) word co-occurrence graph, (2) disjoint graph, (3) complete graph and (4) our graphs. The word co-occurrence graph is created by a simple sliding window method that is not considering the sentence information. Then we learn a disjoint graph in the model by setting threshold $T$ (in the equation \ref{10}) equals to 1, which can lead to none of edges are sampled during structure learning, thus the inter sentence information is completely ignored and the graph only focuses on the intra-sentence information. On the contrary, we set threshold $T$ to 0 to learn a complete graph, where all words inter-sentence are connected to each other. Ideally, this complete graph would have been able to learn relatively global information, however, the sharp increase in the number of edges dissolves the information within sentences, and it cannot learn informative features to perform the task effectively.

The experimental results of these graphs and our graphs are shown in Table \ref{table3}. We note that our graphs perform the best. This suggests that (1) it is useful to use sentence information to construct document-level graph for classification tasks, allowing for word sense disambiguation and capturing synonyms. And (2) sparse structures learned from dynamic contexts in documents can help improve the generalization of document classification. It is worth noting that disjoint and complete graph have different results on different datasets, indicating that each documents owns its characteristics and they need to be learned adaptively according to the goal of document classification.

\begin{table}[t]
\centering
% \resizebox{.95\columnwidth}{!}
\begin{tabular}{l|ccc}
\toprule
\textbf{Graph} & \textbf{R8} & \textbf{R52} & \textbf{Ohsumed} \\ \midrule
WordCooc & 97.20$\pm$0.29 & 93.82$\pm$0.15 & 68.08$\pm$0.32 \\
Disjoint & 97.29$\pm$0.21 & 94.80$\pm$0.20 & 69.72$\pm$0.27 \\
Complete & 97.40$\pm$0.25 & 94.35$\pm$0.10 & 67.57$\pm$0.30 \\
% Complete graph(soft)\\
\midrule
Ours      & 97.76$\pm$0.16 & 95.32$\pm$0.21 & 70.53$\pm$0.30\\
Ours w/ reg & \textbf{97.81$\pm$0.14} & \textbf{95.48$\pm$0.26} & \textbf{70.59$\pm$0.38}\\
\bottomrule
\end{tabular}
\caption{Comparison with different constructions of document-level graphs. (1) WordCooc denotes word co-occurrence graph. (2) Disjoint means a disjoint union of sentence-level subgraphs. (3) Complete graph means disjoint graph with fully connected edges between sentences. (4) Ours graph is constructed by sentence-level subgraphs and learned by sparse structure learning(w/ reg means we add regularization to our model). } 
\label{table3}
\end{table}

\begin{table}[t]
\centering
% \resizebox{.95\columnwidth}{!}
\begin{tabular}{r|ccc}
\toprule
\textbf{$\tau$} & \textbf{R8} & \textbf{R52} & \textbf{Ohsumed} \\ \midrule
0.01 & 97.50$\pm$0.29 & 95.16$\pm$0.18 & \textbf{70.59$\pm$0.38} \\
0.1  & 97.34$\pm$0.13 & \textbf{95.48$\pm$0.26} & 70.21$\pm$0.40 \\
0.2  & 97.44$\pm$0.39 & 95.03$\pm$0.16 & 70.33$\pm$0.32  \\
0.5  &\textbf{97.81$\pm$0.14} & 94.56$\pm$0.33 & 70.34$\pm$0.37 \\
1.0  & 97.35$\pm$0.24 & 95.09$\pm$0.32 & 70.22$\pm$0.29 \\
\bottomrule
\end{tabular}
\caption{Test accuracy with different temperatures $\tau$ for adaptive sampling.}
\label{table4}
\end{table}

\subsection{Adaptive Sampling Analysis}
To obtain a proper sampling temperature for each dataset, we set five temperatures of each dataset for experiments in table \ref{table4}.
During sampling the adaptive neighbors for each nodes via Gumbel-softmax with an adjustable temperature, the smaller the temperature, the more the samples tend to be categorically distributed, which in our model represents the document graph learning to a more sparse structure.
From the results, we can see that different datasets have different proper temperatures, and the Ohsumed dataset reaches the appropriate temperature at a very small value, which meets the same conclusion as the ablation study in the previous section, i.e., the learned sparse structures are adaptive for topic label classification for Ohsumed dataset.
In Appendix, more details for the temperatures of the other datasets and a real case study of visualizing important sparse connections that are learned by our model are provided.

\subsection{Generalization on Imbalanced Unlabeled Data}
\begin{figure}[t]
\centering
\includegraphics[width=1.0\columnwidth]{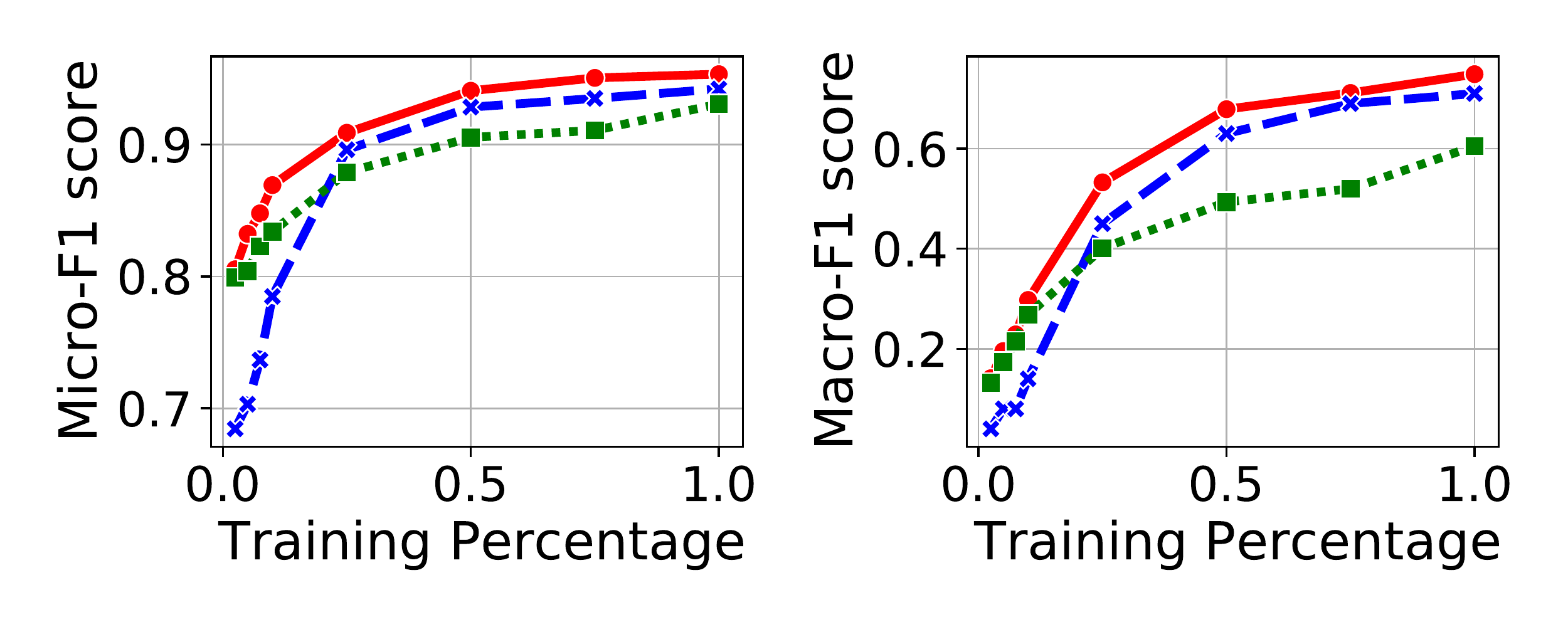} % Reduce the figure size so that it is slightly narrower than the column. Don't use precise values for figure width.This setup will avoid overfull boxes.
\caption{Micro F1 score and Macro F1 score with different percent of training data from 0.025 to 1 on R52 dataset. (Red: Ours; Blue: HyperGAT; Green: TextING)}
\label{fig2}
\end{figure}

To estimate the generalization ability of our model over imbalanced dataset, we use different proportions of training data on R52 dataset to compare with the inductive learning baselines. As the number of training sets decreases, it becomes more challenging to train model on extremely imbalanced datasets. Figure \ref{fig2} shows that all inductive methods achieve improved performance as the number of labeled training data increases.
Our method significantly outperforms other inductive baselines in all cases, indicating that the sparse document structure learned using sentence information allows our model to generalize well even in extremely imbalanced dataset.

\section{Related Works}
\subsection{Document Classification}
Document classification is one of the most fundamental tasks in the field of natural language processing (NLP). Document classification is widely used in many downstream applications, such as spam filtering \cite{wu2020graph}, news classification \cite{liu2018early}, sentiment analysis \cite{medhat2014sentiment}, etc.
An important part of document classification is feature extraction. The traditional approach is to use word-based statistical models to compute features of documents and apply them to downstream classifiers, such as support vector machines \cite{suykens1999least}, naive Bayes \cite{mccallum1998comparison}, random forests \cite{svetnik2003random}, etc.
With the rapid development of deep learning, many feature-based deep learning models have been proposed, such as the study of word representation learning models \cite{mikolov2013distributed, grover2016node2vec}.
Considering the word order in sequences, many models use sequence-based models, including recurrent neural networks (RNNs) \cite{mikolov2010recurrent,liu2016recurrent}, convolutional neural networks (CNNs) \cite{kim-2014-convolutional}.
In addition, there are also models, such as \cite{yang2016hierarchical,8933476}, leverage document hierarchical structures to jointly consider word order-sentence order information.
Since the above models are studied based on sequence data, the dependencies between long sentences may not be taken into account.
Inspired by the semi-supervised GNN proposed by \cite{kipf2017semi}, the study of transforming documents into data structured as graphs and optimizing GNN parameter learning model on the document graphs has rapidly gained attention.

\subsection{GNNs for Document Classification}
The popularity of GNNs have grown rapidly in recent years \cite{kipf2017semi, velivckovic2018graph, hamilton2017inductive,xu2018powerful}. 
In natural language domain, GNN can better capture the non-consecutive phrases and long-distance word dependency in the documents \cite{wu2020comprehensive, li2020survey}. Recent works on GNNs for document classification can be divided into two categories.
One is transductive learning. TextGCN \cite{yao2019graph} first applies GNNs on a whole corpus graph. Huang et al. \cite{huang2019text} build graphs for each document with global shared structure to support effective online learning. TensorGCN \cite{liu2020tensor} jointly learns syntactic, semantic and sequential on a document graph tensor based on a whole corpus. DHTG \cite{wang2020learning} propose a novel hierarchical topic graph to learn an corpus graph with meaningful node embeddings and semantic edges.
While transductive learning models have to be retrained once evaluating an unseen documents, which is unrealistic in the real world.
On the other hand, inductive models can solve this problem. Peng et al. \cite{peng2018large} present a graph-based model to perform hierarchical text classification. TextING \cite{zhang2020every} builds individual word co-occurrence graphs for each document and shows a better generalization performance on unseen documents. 
HyperGAT \cite{ding2020more} proposes novel document-level hypergraphs and inject topic information to obtain high-order semantic information in each document. 
However,the hypergraphs with pre-defined latent topics lacks local syntactic information. 
Our proposed model makes first attempt to leverage sequence information to construct novel document-level graphs that can jointly aggregate local syntactic and global semantic information to learn fine-grained word representations for autonomous inductive document classification. 

\section{Conclusion}

In the real world, each document has its own rich sentence structure, where the intra-sentence context contains local information, and the inter-sentence context captures long-range word dependencies. To this end, we construct a novel trainable document-level graph to jointly capture local and global contextual information. We propose a sparse structure learning via GNNs to refine the structure of the graph with learned dynamic context dependencies. Experimental results show that our proposed approach of combining and learning local and global information is effective for inductive document classification.

\section{Training Details}
More detailed information for model training is as follows. The batch size is chosen from \{16, 64, 128, 256\}. The initial node embedding is obtained using  pre-trained GloVe \cite{pennington2014glove} with dimension 300 and the hidden node dimension is chosen from \{96, 256, 512\}.  For each dataset, we select learning rate from \{1e-4, 5e-4, 1e-3\} and dropout rate from \{0.0, 0.1, 0.3, 0.5, 0.7, 0.9, 1.0\}. we select the point with the highest validation in 200 epochs to estimate the performance of test set. We use Adam (Kingma and Ba 2014) to optimize the model and We use PyTorch (Paszke et al. 2019) and Pytorch Geometric \cite{fey2019fast} to implement our architecture. For the TexTING \cite{zhang2020every} and HyperGAT \cite{ding2020more} baselines, we use the same dataset and hyperparameters provided by the authors for a fair comparison.

\section{Preprocessing Details}
For the quantitative assessment, all data preprocessing is the same as in \cite{kim-2014-convolutional, yao2019graph}. For constructing word co-occurrence graphs for datasets, the size of window sliding on the sentences is set to 3. On this basis, to satisfy the study conditions of the model, we segment the sentences with NLTK \cite{bird2006nltk}.
However, for the R8 and R52 datasets, they are only provided by the preprocessed version, lack punctuation, and do not have explicit sample names. Since we use documents with sentence segmentation information to construct the graphs, we re-extract the data from the original Reuters-21578 dataset. We first filter the samples with only one label in the 115 categories dataset from the Reuters source data, and then select all corresponding categories in the R52 data in the training set and the test set, respectively. After filtering, we find that the training set samples in the test set are consistent with the dataset provided by \cite{yao2019graph}. The corresponding datasets are provided in the Supplementary file.

\section{Document Classification Performance}
Table \ref{textgraph} shows the results of different constructions of the document graphs on the other two data sets. From the table, we know that these results are consistent with the elaboration in the text. In all datasets, the sparse graph learned from our model using dynamic contextual information is more effective than all other static word co-occurrence graphs, indicating that it has a positive effect on document classification.
Table \ref{text temperature} shows the experimental results for the other two data using different temperatures to estimate the optimal performance. It can be seen that the 20NG and MR data sets perform best at a temperature of 0.5. This is different from the optimal temperature for the three datasets reported in the main paper, again indicating that each dataset has different properties and requires the use of an adapted model to learn the document-specific structure.

\begin{table}[t]
\centering
\begin{tabular}{c|cccccc}
\toprule
\textbf{Graph} & \textbf{MR} & \textbf{20NG}\\ \midrule\midrule
WordCooc graph & 78.42$\pm$0.09 & 84.69$\pm$0.17\\
Disjoint graph & 78.61$\pm$0.12 & 84.92$\pm$0.28\\
Complete graph & 78.77$\pm$0.10 & 83.38$\pm$0.45\\
\midrule
our graph  & 79.74$\pm$0.11 & 85.15$\pm$0.33\\
our graph w/ reg &  \textbf{79.74$\pm$0.19} & \textbf{85.26$\pm$0.28}\\
\bottomrule
\end{tabular}
\caption{Comparison with different constructions of document-level graphs.}
\label{textgraph}
\end{table}
\begin{table}[t]
\centering
\begin{tabular}{c|cccccc}
\toprule
\textbf{Temperature} & \textbf{MR} & \textbf{20NG}\\ \midrule\midrule
0.01 & 79.10$\pm$0.12 & 85.11$\pm$0.11  \\
0.1 & 78.10$\pm$0.14 & 85.01$\pm$0.32  \\
0.2 & 79.66$\pm$0.11 & 84.75$\pm$0.12  \\
0.5 & \textbf{79.74$\pm$0.19} & \textbf{85.26$\pm$0.28}  \\
1.0 & 78.89$\pm$0.13 & 85.10$\pm$0.22  \\
\bottomrule
\end{tabular}
\caption{Test accuracy with different temperatures for adaptive sampling.}
\label{text temperature}
\end{table}

\begin{figure*}[t]
\centering
\includegraphics[width=0.9\textwidth]{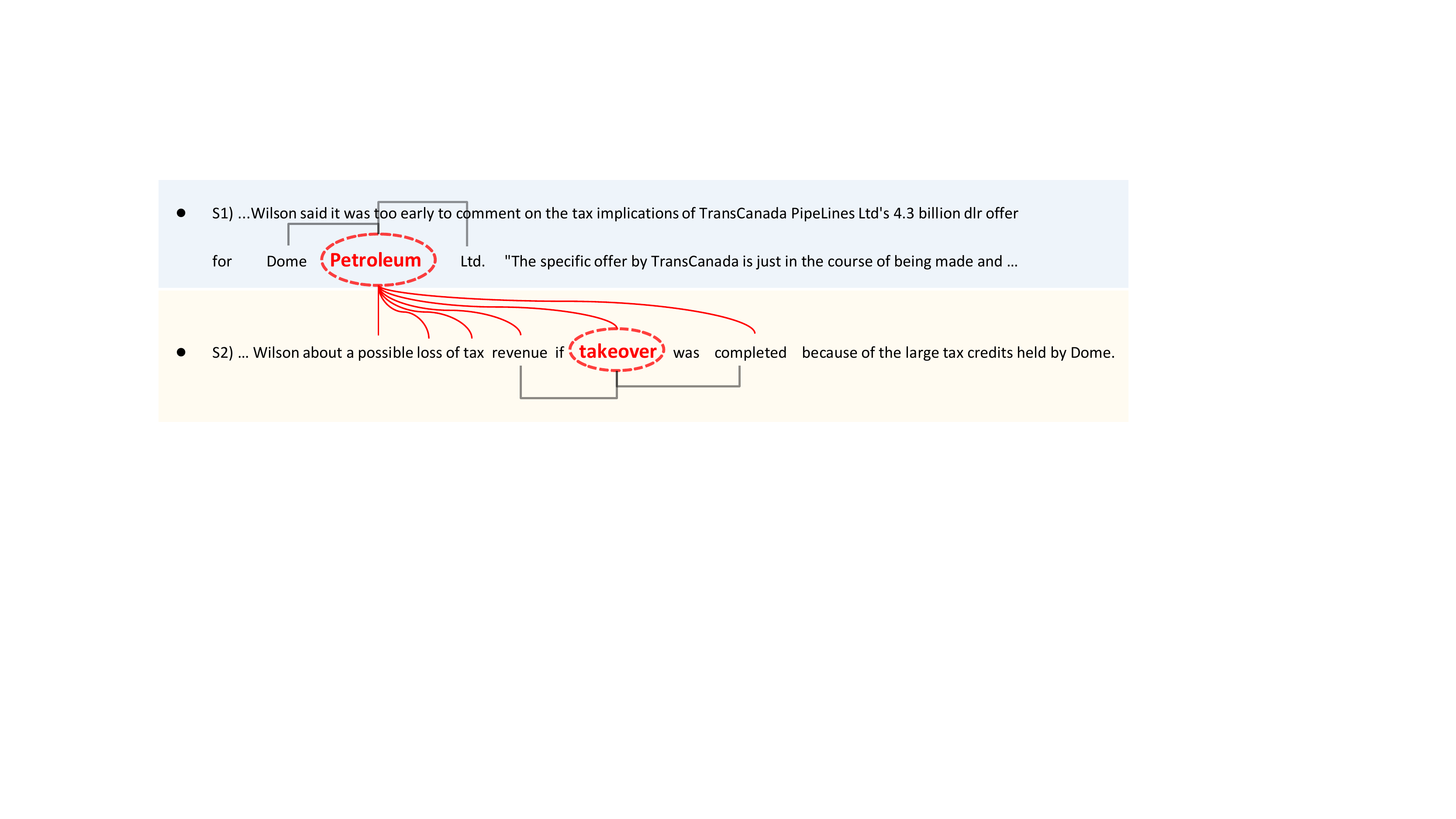} \includegraphics[width=0.9\textwidth]{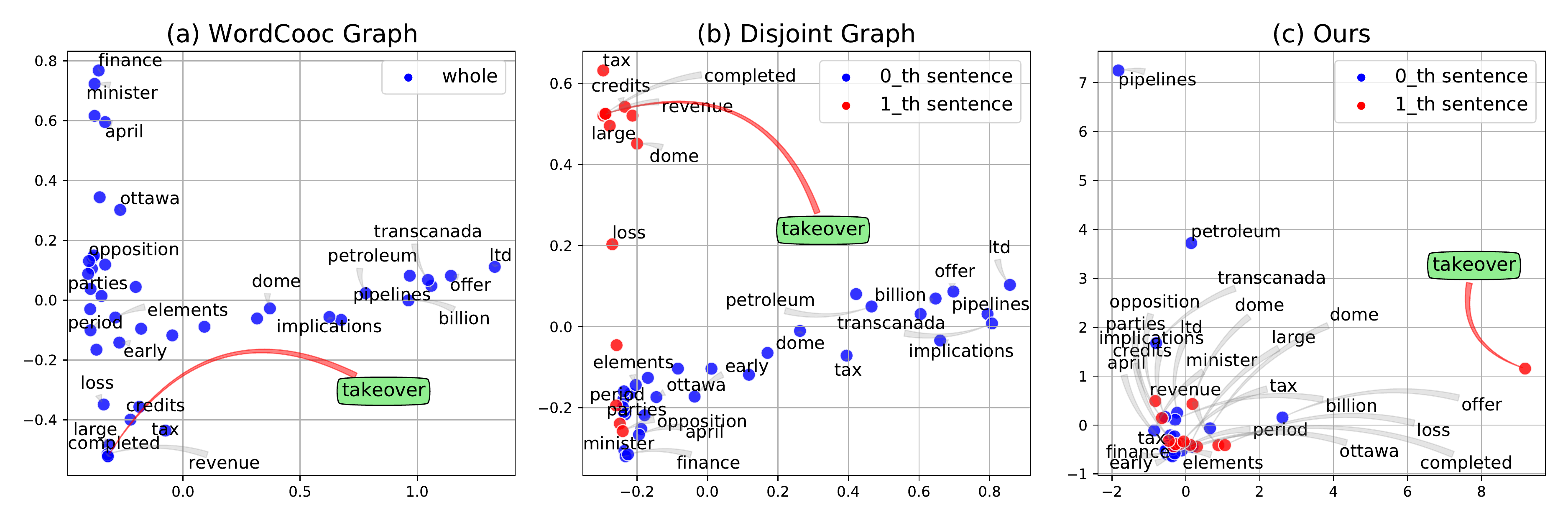}
\caption{A real example document from  R52 dataset. It consists of two sentences ($s1$ and $s2$) with real labels of the ``\textit{acquisition}" class. In the document, we use the black line to indicate the local syntactic connectivity between words and the red line to denote the global semantic connections between sentences learned from our model. The figure below denotes word embeddings from different constructions of document graph: (a) word co-occurrence graph; (b) disjoint graph; (c) our graph.}
\label{fig1}
\end{figure*}

\section{Visualization}
In this subsection, we want to know what important sparse connections are learned by our model.
We extract a real example text from R52 dataset, as shown in Figure \ref{fig1}. The document consists of two sentences ($s1$ and $s2$) belonging to a real label ``\textit{acquisition}".
We note that in our model, the words ``\textit{takeover}" and ``\textit{petrolum}" learn not only their own local syntactic information separately, but also global semantic connections between the two of them in conjunction with dynamic contextual dependencies.

Models using word co-occurrence graphs and disjoint graphs incorrectly predict this sample as ``\textit{crude}" and ``\textit{earn}", respectively. 
Word embeddings learned from the last GNN layer of each method are plotted by PCA as a two-dimensional visualization as shown in Figure \ref{fig1} below.
Our observations are as follows:\\
(1) The static word co-occurrence graph mispredict to ``\textit{crude}" that is related to ``\textit{petrolum}". This may be due to the fact that not many repeated anchor words in the two sentences of this sample, so the ``\textit{petrolum}" word vector with higher centrality is more prominent in the whole graph.\\
(2) We found a tendency of the words between sentences in disjoint graph to be separated from each other in the latent space. Since it lacks information between sentences, it is also difficult to capture keywords that occur with small frequency, such as ``\textit{takeover}".\\
(3) Our model learns global semantic information by learning local syntactic information and then combining it with dynamic contexts. Instead of simply interpreting ``\textit{petrolum}" , we combine global information and correlate ``\textit{petrolum}" information with ``\textit{takeover}" to achieve correct classification of ``\textit{acquisition}" categories, which indicate that the sparse structure learned from our model helps inductive document classification.

% \begin{table*}[t]
% \centering
% \begin{tabular}{c|cccccc}
% \toprule
% \textbf{dataset} & \textbf{Hidden dimension} & \textbf{Learning rate} & \textbf{Dropout} & \textbf{\#Layers} & \textbf{batch size} \\ \midrule
% MR \\
% R8 \\
% R52 \\
% Ohsumed \\
% 20NG \\
% \bottomrule
% \end{tabular}
% %}
% \caption{Hyperparameter settings for each dataset.}
% \label{hyperparameters}
% \end{table*}

\section{Acknowledgements}
This research was supported by Institute of Information \& communications Technology Planning \& Evaluation (IITP) grant funded by the Korea government(MSIT) [NO.2021-0-01343, Artificial Intelligence Graduate School Program (Seoul National University)] (IITP-2021-0-01343) 
and by the Bio \& Medical Technology Development Program of the National Research Foundation (NRF) funded by the Ministry of Science \& ICT (NRF-2019M3E5D4065965) 
and by Basic Science Research Program through the National Research Foundation of Korea (NRF) funded by the Ministry of Education (NRF-2021R1A6A3A01086898).

\bibliography{aaai22}
\end{document}